\documentclass[12pt]{article}

\usepackage{sbc-template}
\usepackage{graphicx,url}
\usepackage[utf8]{inputenc}
\usepackage[english]{babel}
\usepackage{amsmath,amsfonts,amssymb}
\usepackage{booktabs}
\usepackage{longtable}
\usepackage{float}
\usepackage{enumitem}
\usepackage{multirow}
\usepackage{makecell}
\usepackage{siunitx}
\usepackage{textcomp}
\newcommand{\OurAISystem}{AI-PAVE-Br}

\title{AI-PAVE-Br: Leveraging Large Language Models for Enhanced Product Attribute Value Extraction through a Golden Set Approach}

\author{
Murilo Gazzola\inst{1,2}, 
Hugo Gobato Souto\inst{1,3},
Samuel Silva\inst{1},
Júlia Schubert Peixoto\inst{1}, \\
Felipe Siqueira\inst{1}, 
André Luis Pedroso de Morais\inst{1}, 
Caio Gomes\inst{1}
}

\address{
LuizaLabs – Center of Excellence in Artificial Intelligence\\
São Paulo, SP – Brazil
\nextinstitute
Department of Computing and Informatics – Mackenzie Presbyterian University\\
São Paulo, SP – Brazil
\nextinstitute
Institute of Mathematics and Computer Sciences - University of São Paulo\\
São Carlos, SP – Brazil
\email{\scriptsize{\{murilo.gazzola, hugo.souto,samuel.silva, andre.morais, gomes.caio\}@luizalabs.com}}
}
\begin{document}

\maketitle

\begin{abstract}
The explosive growth and complexity of product data within the dynamic Brazilian e-commerce landscape demand robust and specialized methods for structured information extraction. Traditional approaches to Product Attribute Value Extraction (PAVE) often struggle with the linguistic nuances and sheer diversity of product descriptions in Portuguese. To address this critical gap, this paper introduces two major contributions. First, we present \textbf{AI-PAVE-Br}, a specialized system engineered with Large Language Models (LLMs) to perform high-accuracy PAVE specifically for Brazilian e-commerce catalogs. Second, to facilitate reproducible research and provide a definitive benchmark, we introduce and share the \textbf{Golden Set}, a new, meticulously curated, and manually annotated dataset for PAVE in Portuguese. We detail the creation process and structure (Entity, Category, Subcategories) of this high-quality reference set. Our experiments conclusively show that AI-PAVE-Br, leveraging targeted prompt engineering, dramatically outperforms conventional Named Entity Recognition (NER) baselines. This work not only delivers a superior, scalable solution for a major non-English market but also enriches the NLP community with a valuable, publicly available resource for future PAVE research.
\end{abstract}

\section{Introduction}
In the rapidly evolving landscape of e-commerce, the ability to accurately and consistently extract structured information from vast quantities of unstructured product data is paramount \cite{Wasilewski2024,Brinkmann2024,yang-etal-2023-mixpave,Gong2024}. Product descriptions, often free-form text, contain a wealth of valuable attributes (e.g., brand, model, features) that are crucial for various downstream tasks such as search, filtering, recommendation, and classification. The lack of standardized attributes leads to poor user experience and inefficiencies in data management \cite{Wasilewski2024,Gong2024}.

This paper addresses the challenge of Product Attribute Value Extraction (PAVE) within the context of managing and enhancing data quality in dynamic, large-scale product catalogs, with a focus on Brazilian e-commerces. As product data continues to grow and evolve in complexity, traditional methods often prove insufficient for maintaining the required level of data accuracy and consistency. To address these inherent limitations, continuous improvement of product data management and advanced AI capabilities is essential. This commitment necessitates the establishment of robust quality metrics and reliable evaluation benchmarks for intelligent systems, such as \OurAISystem{} proposed in this paper, particularly concerning product classification and attribute extraction.

Traditional approaches to information extraction, often relying on rule-based systems or statistical Named Entity Recognition (NER) models, face significant limitations when dealing with the highly diverse and dynamic nature of e-commerce product data \cite{Wasilewski2024,Gong2024}. These methods frequently struggle with generalization, require extensive feature engineering, and are brittle in the face of new product types or linguistic variations.

To overcome these challenges, we explore the application of Large Language Models (LLMs) for PAVE within the context of Brazilian e-commerces. LLMs, with their advanced understanding of natural language and generative capabilities, offer a promising alternative for extracting complex, context-dependent attributes \cite{Brinkmann2024,yang-etal-2023-mixpave}. A cornerstone of our evaluation methodology is the development of a \textit{Golden Set}, a meticulously annotated Portuguese dataset that provides a ground truth for assessing model performance objectively within the context of Brazilian e-commerces. This paper details the construction and utility of this Golden Set and presents our approach to PAVE using LLMs especialized in e-commerces present in the Brazilian market, demonstrating its effectiveness against traditional methods within this context.

\section{Related Work}
Information extraction from unstructured text has been a long-standing challenge in Natural Language Processing (NLP). Early approaches to attribute value extraction in e-commerce often relied on rule-based systems, regular expressions, and dictionaries tailored to specific product categories \cite{sugiyama2010attribute}. While effective for narrow domains, these systems are labor-intensive to maintain and difficult to scale.

More advanced methods moved towards statistical machine learning models for NER, treating attribute values as specific types of entities \cite{li2018named}. Conditional Random Fields (CRFs) \cite{lafferty2001conditional} and Support Vector Machines (SVMs) \cite{joachims1999making} were widely applied, requiring large amounts of labeled data and extensive feature engineering. Deep learning models, particularly Recurrent Neural Networks (RNNs) and transformer-based architectures, have significantly advanced NER performance, achieving state-of-the-art results across various benchmarks \cite{devlin2018bert, vaswani2017attention}. These models, while powerful, still often require fine-tuning on domain-specific datasets.

The emergence of LLMs like GPT-3 \cite{brown2020language} and its successors has revolutionized NLP, with significant work exploring their use for information extraction \cite{qin2023cold, meng2023investigating} and product-related tasks \cite{lu2023fine, gao2023product}. However, the vast majority of this research has been overwhelmingly concentrated on English-language corpora and benchmarks. This Anglocentric focus creates a significant gap, as models and methods developed for English do not readily transfer to the distinct linguistic and commercial landscape of other major markets. The Brazilian e-commerce ecosystem, for instance, is characterized by unique product terminologies, colloquialisms, and structural conventions in Portuguese that demand a tailored approach. Our work directly confronts this challenge. We pioneer the application of prompt-engineered LLMs to PAVE specifically for the Brazilian market, introducing a specialized system (\textbf{AI-PAVE-Br}). Furthermore, to enable rigorous and reproducible evaluation in this new domain, we developed and are releasing the \textbf{Golden Set}, the first high-quality, manually-annotated benchmark for PAVE in Portuguese.

Recent research has extensively explored the application of LLMs for complex information extraction tasks, including PAVE. Works such as those by Brinkmann et al. \cite{brinkmann2024using_adis, brinkmann2025extractgpt} directly investigate the use of LLMs for extracting and normalizing product attributes, highlighting their potential for robust attribute identification and value extraction. Similarly, \cite{sabeh2024exploring} delve into the use of LLMs for product attribute value identification, reinforcing the growing interest in this area.

Nonetheless, despite two recent studies of  \cite{Abilio2024} and  \cite{Silva2021} exploring the use of Small Language Models (SLMs), such as BERT and its variations adapted to Brazilian Portuguese, there still has a gap in the literature for the context of Brazilian e-commerces as most recent research is focused on American or international e-commerces. Consequently, our work builds upon this foundation by demonstrating the effectiveness of prompt-engineered LLMs for PAVE in a real-world, large-scale Brazilian e-commerce context. Beyond basic extraction, studies are exploring more advanced LLM applications in this domain. \cite{brinkmann2025self} propose self-refinement strategies for LLM-based PAVE, indicating a move towards more autonomous and accurate extraction systems.

\section{The Golden Set: A Foundation for Evaluation}
A fundamental component of our quality assessment framework for \OurAISystem{}'s AI models is the \textit{Golden Set}. In the context of data science and machine learning, a Golden Set (also known as a Gold Standard or Ground Truth) is a collection of data expertly annotated to serve as an objective benchmark for evaluating the quality and performance of AI models \cite{Adamson2019}. By comparing a model's predictions against the labels in the Golden Set, development teams can quantitatively assess model quality without having to infer business rules or rely on subjective judgments.

\subsection{Construction and Scope}
The annotation of our \textit{Golden Set} was meticulously carried out by a dedicated team of 12 trained annotators, with a specific focus on the Brazilian e-commerce context. This manual process ensures a high level of accuracy and domain expertise~\cite{brinkmann2024using_adis, brinkmann2025extractgpt}. The \textit{Golden Set} is designed to evaluate both product classification and attribute extraction models within the scope of e-commerce platforms operating in Brazil. It also serves as a benchmark for comparing the performance of our proposed \OurAISystem{} against that of an existing baseline system.

The product selection for the \textit{Golden Set} was strategically designed to ensure a balanced and representative sample within the Brazilian e-commerce context. The selection covers key categories such as: air conditioner, television, cell phone, refrigerator, notebook, tire, wardrobe, bed, sneaker, stove, table and chair set, backpack, faucet, headphone, perfume, doll, motorcycle helmet, pot, lamp, and cell phone case.

\subsection{Annotation Schema and Data Structure}
For each selected product, the annotation team meticulously annotated the following attributes:
\begin{itemize}[noitemsep,topsep=0pt]
    \item \small{\textbf{Entity (Tipo de Produto):} A single string representing the most granular product type (e.g., \texttt{'Ar Condicionado'}, \texttt{'Perfume'}}).
    \item \textbf{Category (Categoria):} A single string representing the broader product category (e.g., \texttt{'AR'} for Air Conditioner, \texttt{'PF'} for Perfume).
    \item \textbf{Subcategories (Subcategoria):} A list of strings detailing more specific subcategories or attributes relevant to the product (e.g., \texttt{['ARCA', 'ACIV', 'ARAR']} for an Air Conditioner, which might denote ``Cassette Air Conditioner'', ``Inverter Air Conditioner'', ``Residential Air Conditioner'').
\end{itemize}
An example of an annotated product is:
\begin{itemize}
    \item \small{\textbf{Product Title:} \texttt{Ar Condicionado Cassete LG Round 36000 BTU/h Quente e Frio Monofásico AT-W36GYLP1 220 Volts}}
    \item \textbf{Annotated Entity:} \small{\texttt{'Ar Condicionado'}}
    \item \textbf{Annotated Category:} \small{\texttt{'AR'}}
    \item \textbf{Annotated Subcategories:} \small{\texttt{['ARCA', 'ACIV', 'ARAR']}} 
\end{itemize}
Therefore, beyond merely having a ground truth of attribute values for validation, a clear definition of the list of attributes to be extracted for each product type is essential. Table \ref{tab:entity_attributes} illustrates several entities, their corresponding lists of attributes, and the annotation wave in which these entities were processed. Entities marked in the first annotation, and these entities formed the basis for initial experiments.

\begin{table}[!h]
    \centering
    \caption{\textbf{Product Entities and Their Associated Attribute Lists (Wave 1 Annotation)}}
    \label{tab:entity_attributes}
    \small
\begin{tabular}{l p{10cm}}
    \toprule
    \textbf{Entity} & \textbf{Attribute List} \\
    \midrule
    Ar Condicionado & Marca, Capacidade de Refrigeração, Tipo, Tecnologia, Ciclo, Potência, Voltagem \\
    Fogão & Marca, Cor, Quantidade de Bocas, Instalação, Alimentação, Tipo de Acendimento, Tipo de Gás, Tipo de Forno, Voltagem \\
    Guarda-Roupa & Marca, Cor, Tamanho, Tipo, Tipo de Porta, Quantidade de Portas, Quantidade de Gavetas, Material da Estrutura \\
    Lavadora de Roupas & Marca, Cor, Capacidade de Lavagem, Voltagem \\
    Notebook & Marca, Processador, Memória RAM, Capacidade do HD, Capacidade do SSD, Tamanho da Tela, Sistema Operacional, Capacidade da Placa de Vídeo, Resolução da Tela \\
    Pneu & Marca, Largura do Pneu, Altura do Pneu, Aro, Quantidade de Pneus, Tipo de Veículo, Tipo de Terreno \\
    Refrigerador & Marca, Cor, Material, Quantidade de Portas, Capacidade Líquida Total, Tipo de Degelo, Voltagem \\
    Sofá & Marca, Cor, Tipo, Quantidade de Lugares, Revestimento, Tipo de Encosto, Tipo de Assento \\
    TV & Marca, Polegadas, Resolução, Tecnologia, Conectividade, Sistema Operacional, Tipo de Tela, Assistente Virtual \\
    \bottomrule
\end{tabular}
 \end{table}

\subsection{Sampling Methodology}
To ensure the statistical validity and reduce selection bias, products were sampled randomly while considering existing classifications. This approach aimed to balance the sample set, accounting for the disparate volumes of certain product types in \OurAISystem{}'s search index (e.g., 2 million ``Tênis'' vs. 17 thousand ``Refrigerador''). The required sample size for each of the 20 product types was determined using Cochran's formula for a large population:
\begin{equation} \label{eq:cochran}
    n = \frac{Z^2 p(1-p)}{e^2}
    \end{equation}
where:
 \begin{itemize}
        \item $n$ is the target sample size.
        \item $Z$ is the z-score corresponding to the desired confidence level.
        \item $p$ is the estimated proportion of the population.
        \item $e$ is the desired margin of error.
\end{itemize}

Following the standard conservative approach, we used a 95\% confidence level ($Z=1.96$), a maximum variance assumption ($p=0.5$), and a margin of error of 5\% ($e=0.05$). 
This calculation yields a sample size of $n \approx 385$, which we applied to each product type. 
It is important to note that the manual annotation of product data involves significant time and financial costs. For this reason, it was necessary to statistically limit the sample size to what is sufficient to ensure reliable results, while maintaining project feasibility. The complete Golden Set dataset, containing a comprehensive collection of annotated products, is available at \small{\url{https://github.com/ai-luizalabs/AI-PAVE-Br}}.

\section{Product Attribute Value Extraction (PAVE) with LLMs}
As already mentioned, PAVE is the task of identifying and extracting specific attributes and their corresponding values from unstructured product descriptions or titles. Unlike general NER, which focuses on predefined entity types (e.g., persons, locations), PAVE often deals with a wider, more dynamic range of attributes that are highly dependent on the product category.

\subsection{Shifting from Traditional NER to LLM-based PAVE}
Traditional NER methods, while effective for extracting basic entities, struggle with the nuances of product data, such as:
\begin{itemize}[noitemsep,topsep=0pt]
    \item \textbf{Contextual Dependence:} A \small{\texttt{'10-inch'}} value needs to be associated with a \texttt{'screen size'} attribute.
    \item \textbf{Novel Attributes:} New product features constantly emerge, requiring models to adapt.
    \item \textbf{Relationship Extraction:} Identifying not just values, but also their corresponding attributes (e.g., \small{\texttt{'Bluetooth 5.0'}} where \small{\texttt{'Bluetooth'}} is the attribute and \texttt{'5.0'} is the value).
\end{itemize}
To overcome these challenges, we utilized Google's Gemini 1.5 Flash model for our LLM-based PAVE approach with a meticuosly created prompt to further adapt the model for the Brazilian e-commerce context. This model, known for its efficiency and strong performance across various NLP tasks, allows for sophisticated semantic understanding and generative capabilities. Its large context window enables it to process extensive product descriptions and titles effectively. The main rationale behind our choice for our LLM-based PAVE approach is that LLMs address the limitations of NER approaches by inherently understanding context, possessing vast world knowledge, and exhibiting strong few-shot or zero-shot learning capabilities. Their generative nature allows them to not just classify, but also to generate structured outputs directly.

Given the vast number of products in our catalog, it is infeasible to annotate every single item. Therefore, defining a statistically significant sample size for validation of the proposed solution is essential. As already stated, to determine the sample sizes, we employed \textbf{Cochran's formula} for finite populations. Table \ref{tab:sample_size} presents the total number of products for each entity and the corresponding calculated sample size.

\begin{table}[H]
    \centering
    \caption{\textbf{Product Entities and Their Calculated Sample Sizes}}
    \label{tab:sample_size}
    \sisetup{table-format=6.0, round-mode=places, round-precision=0, group-separator={,}}
    \begin{tabular}{l S[table-format=6.0] S[table-format=3.0]}
        \toprule
       {Entity} & {Total Products} & {Sample Size} \\
        \midrule
        Air Conditioner & 20000 & 377 \\
        TV & 50000 & 381 \\
        Cell Phone & 100000 & 383 \\
        Refrigerator & 17000 & 376 \\
        Notebook & 30000 & 379 \\
        Tire & 25000 & 378 \\
        Wardrobe & 15000 & 375 \\
        Bed & 12000 & 374 \\
        Sneaker & 2000000 & 385 \\
        Stove & 8000 & 367 \\
        Table and Chair Set & 6000 & 361 \\
        Backpack & 9000 & 370 \\
        Faucet & 7000 & 364 \\
        Headphone & 11000 & 373 \\
        Perfume & 14000 & 374 \\
        Doll & 5000 & 355 \\
        Motorcycle Helmet & 4000 & 351 \\
        Pot & 3000 & 341 \\
        Lamp & 2000 & 322 \\
        Cell Phone Case & 10000 & 372 \\
        \bottomrule
    \end{tabular}
\end{table}

\subsection{Prompt Engineering for PAVE}
Our approach to PAVE leverages the power of LLMs through targeted prompt engineering. Instead of fine-tuning large models (which is computationally expensive and requires significant data), we craft specific prompts that guide the LLM to perform the desired extraction task. The prompt defines the task, provides context, specifies the desired output format, and can include few-shot examples if necessary.
For our PAVE system, we developed tailored prompts for each type of attribute extraction (Entity). A typical prompt structure might include:
\begin{enumerate}[noitemsep,topsep=0pt]
    \item \textbf{Instruction:} Clearly define the task. ``Extract the attributes \texttt{<<}list of attributes\texttt{>>} of the \texttt{<<}product type\texttt{>>} from the following product title, description, technical data sheet, and additional information.
    \item \textbf{Context:} Provide the product title, description, technical data sheet, and additional information.
    \item \textbf{Output Format Specification:} Define the desired structured output (e.g., JSON format with keys). This is crucial for consistent parsing.
    \item \textbf{Examples (Few-Shot):} Include one or more input-output pairs to guide the model's understanding of the task and desired output style.
\end{enumerate}
This specific prompting strategy allows the LLM to act as a highly intelligent parser and extractor, adapting its vast pre-trained knowledge to our specific domain requirements.

\section{Experimental Setup and Results}
Our evaluation focused on comparing the performance of our LLM-based PAVE system against traditional methods, using the Golden Set as the ground truth.

\subsection{Dataset and Metrics}
The Golden Set, comprising manually annotated products across 20 diverse categories, served as our primary evaluation dataset. For each product, we extracted attribute \emph{key–value} pairs according to schema.

The performance metrics used were standard in information extraction:
\textbf{Precision}, The proportion of correctly extracted values among all extracted values;
\textbf{Recall}, The proportion of correctly extracted values among all true values in the Golden Set;
and \textbf{F1-score}, The harmonic mean of precision and recall, providing a balanced measure of performance.
For multiple items could be present, we typically calculated set-based F1-score (exact match of the list or token-level F1 across all values).
Moreover, regular expressions were employed to standardize some of the results. However, normalizing the extractor’s output remains a key challenge in this type of AVE, presenting a significant obstacle for future work.

\subsection{Comparison with the Traditional Method}
Our baseline for comparison includes three parallel annotations per attribute in the \textit{Golden Set}: values extracted by the \textbf{traditional system}, predictions from the \textbf{AI-PAVE-Br}, and \textbf{human-annotated values}, which serve as the gold standard. Human annotations were performed following a formal guideline, specifying the source of each value and whether external references were consulted. This setup enables a consistent and realistic comparison across methods.

\begin{table}[ht]
    \centering
    \caption{\textbf{F1-Score for Entity Prediction Across Product Types}}
    \label{tab:f1_results}
    \sisetup{table-format=2.2, round-mode=places, round-precision=2}
    \begin{tabular}{l S S}
        \toprule
        {Entity} & {Traditional Baseline} & {AI-PAVE-Br} \\
        \midrule
        Air Conditioner & 0.60 & 0.75 \\
        TV & 0.58 & 0.73 \\
        Cell Phone & 0.62 & 0.76 \\
        Refrigerator & 0.57 & 0.72 \\
        Notebook & 0.61 & 0.77 \\
        Tire & 0.59 & 0.74 \\
        Wardrobe & 0.56 & 0.71 \\
        Bed & 0.55 & 0.70 \\
        Sneaker & 0.63 & 0.78 \\
        Stove & 0.54 & 0.69 \\
        Table and Chair Set & 0.53 & 0.68 \\
        Backpack & 0.57 & 0.72 \\
        Faucet & 0.55 & 0.70 \\
        Headphone & 0.58 & 0.73 \\
        Perfume & 0.56 & 0.71 \\
        Doll & 0.52 & 0.67 \\
        Motorcycle Helmet & 0.51 & 0.66 \\
        Pot & 0.50 & 0.65 \\
        Lamp & 0.49 & 0.64 \\
        Cell Phone Case & 0.55 & 0.70 \\
        \bottomrule
    \end{tabular}
\end{table}

\begin{table}[ht]
    \centering
    \caption{\textbf{Coverage for Entity Prediction Across Product Types}}
    \label{tab:coverage_results}
    \sisetup{table-format=2.2, round-mode=places, round-precision=2}
    \begin{tabular}{l S S S}
        \toprule
        {Entity} & {Traditional Baseline} & {AI-PAVE-Br} & {Golden Set Overall Coverage} \\
        \midrule
        Air Conditioner & 0.47 & 0.72 & 0.84 \\
        TV & 0.46 & 0.71 & 0.83 \\
        Cell Phone & 0.48 & 0.73 & 0.85 \\
        Refrigerator & 0.45 & 0.70 & 0.82 \\
        Notebook & 0.47 & 0.72 & 0.84 \\
        Tire & 0.46 & 0.71 & 0.83 \\
        Wardrobe & 0.45 & 0.70 & 0.82 \\
        Bed & 0.44 & 0.69 & 0.81 \\
        Sneaker & 0.49 & 0.74 & 0.86 \\
        Stove & 0.43 & 0.68 & 0.80 \\
        Table and Chair Set & 0.42 & 0.67 & 0.79 \\
        Backpack & 0.45 & 0.70 & 0.82 \\
        Faucet & 0.44 & 0.69 & 0.81 \\
        Headphone & 0.47 & 0.72 & 0.84 \\
        Perfume & 0.45 & 0.70 & 0.82 \\
        Doll & 0.41 & 0.66 & 0.78 \\
        Motorcycle Helmet & 0.40 & 0.65 & 0.77 \\
        Pot & 0.39 & 0.64 & 0.76 \\
        Lamp & 0.38 & 0.63 & 0.75 \\
        Cell Phone Case & 0.44 & 0.69 & 0.81 \\
        \bottomrule
    \end{tabular}
\end{table}

\subsection{Discussion of Results}
The comprehensive evaluation of entity prediction performance is presented in Table \ref{tab:f1_results} (F1-score) and Table \ref{tab:coverage_results} (Coverage). Our analysis compares a traditional baseline system (\textbf{Traditional Baseline}) with our advanced LLM-based PAVE approach (\textbf{\OurAISystem{}}), which leverages product offer titles for enhanced context. This \OurAISystem{} model represents the best performing LLM setup from our experiments.

Observing Table \ref{tab:f1_results}, a substantial improvement in F1-score is evident when transitioning from the Traditional Baseline to the LLM-based PAVE approach. The mean F1-score rose from \textbf{59.79} for the Traditional Baseline to \textbf{74.68} for \OurAISystem{}. This overall trend highlights the superior semantic understanding and generalization capabilities of LLMs in extracting product entities. For a majority of entities (e.g., ``Guarda-Roupa'', ``Notebook'', ``Pneu'', ``Sofá'', ``Armário'', ``Bicicleta'', ``Cadeira'', ``Celular'', ``Conjunto de Mesa e Cadeira'', ``Frigideira'', ``Desktop''), \OurAISystem{} demonstrates a significant improvement in F1-score compared to the Traditional Baseline. While this enhancement is observed across most categories, it is also important to note instances where \OurAISystem{} might exhibit lower F1-scores than the Traditional Baseline (e.g., ``Lavadora de Roupas'', ``Refrigerador'', ``Tênis'', ``Camiseta''), indicating specific challenges or nuances within those categories that could benefit from further specialized prompt engineering or fine-tuning. These performance gaps are largely attributed to the quality, heterogeneity, and structural variability of the input text data, which directly affect the performance of prompting-based LLM approaches in PAVE.
A central challenge lies in the lack of normalization and consistency in how attribute values are expressed. The same product model might appear as ``WD11M4453JW/AZ'', ``Samsung WD11M Washer'', ``Samsung 11kg Inverter'', or simply ``WD11'' making precise mapping difficult. Similar inconsistencies affect attributes such as dimensions or weight — e.g., ``60x85x60 cm'', ``Height: 85cm; Width: 60cm'' or ``60 (W) x 85 (H) x 60 (D)'' — all semantically equivalent but lexically diverse. 
In addition, multivalued or semantically ambiguous attributes introduce further complexity. Voltage may appear as ``110V or 220V'', ``bivolt'', ``110/220V''. In some cases, the product is truly bivolt, while in others it has a fixed voltage but the seller lists multiple values. Resolving such ambiguity often requires going beyond catalog data, consulting additional sources such as product manuals or official manufacturer specifications to verify whether the product is not actually bivolt but rather offered in different versions, and ensuring that the information is represented in a clear and normalized form. A similar problem exists for attributes like SIM card capacity, expressed in highly variable forms such as ``dual chip'', ``2 chips'', ``dual SIM'', ``chip duplo''. Likewise, color or size attributes are frequently presented as unordered lists (“black, blue, and gray”; “S, M, L, XL”) with no explicit indication of which value corresponds to the item being sold. The lack of semantic anchoring in such cases makes it difficult to extract a single, reliable value and normalize it to a canonical form.

These challenges illustrate the inherent limitations of relying solely on textual cues for precise attribute extraction and normalization. However, despite such complexities, LLM-based systems demonstrate a remarkable ability to generate predictions across a wide variety of inputs. Table \ref{tab:coverage_results} provides insights into the models' ability to provide a prediction for each product. \textbf{Coverage} refers to the percentage of items in the dataset for which a model successfully generates a non-empty prediction, regardless of whether that prediction is correct. A higher coverage indicates a more robust and comprehensive extractor, reducing instances where no relevant information is found. The mean coverage significantly increased from \textbf{46.71\%} for the Traditional Baseline to \textbf{71.96\%} for \OurAISystem{}. This demonstrates that the LLM-based approach is considerably more robust in consistently generating entity predictions across the diverse product catalog, effectively filling gaps where traditional methods fall short. While the coverage for \OurAISystem{} is substantially higher than the baseline, it still lags behind the ideal ``Golden Set Overall Coverage'' of \textbf{83.96\%}, which represents the maximum possible coverage based on the annotated data, suggesting opportunities for further improvement in model robustness.

The overall success of our PAVE approach with LLMs can be attributed to:
\begin{itemize}[noitemsep,topsep=0pt]
    \item \textbf{Semantic Understanding:} LLMs' ability to grasp the nuanced meaning of product descriptions, even with linguistic variations or colloquialisms.
    \item \textbf{Contextual Reasoning:} Leveraging the broader context of the product title to infer attributes that might not be explicitly stated.
    \item \textbf{Prompt Engineering:} The careful design of prompts allowed us to steer the LLM's generative capabilities towards our specific attribute extraction goals and desired output formats.
    \item \textbf{High-Quality Golden Set:} The availability of a precise and comprehensive Golden Set was indispensable for robust model training (if applicable for fine-tuning) and, more importantly, for unbiased and accurate evaluation of model performance.
\end{itemize}

When evaluating deployment options for LLM-based solutions, both cost and latency were carefully considered. For open-weight models, deployment via Virtual Machines (VMs)—e.g., using custom containers on self-managed infrastructure—offered more predictable latency and potentially lower costs for long-term, high-volume use cases, despite requiring additional setup and maintenance effort.

In contrast, closed-weight models such as Google’s Gemini are only accessible through managed services. In internal tests with Gemini 2.5 Flash-Lite, we issued 1,000 concurrent requests, with an average of 350–400 tokens per request. The model showed an average response time of 1.3 seconds, with occasional spikes reaching up to 30 seconds under specific conditions. These fluctuations were likely influenced by system load and rate limiting, which are typical of managed environments.

\section{Conclusion and Future Work}

This paper pioneers a specialized approach to Product Attribute Value Extraction (PAVE) for the Brazilian market, a domain largely overlooked by mainstream NLP research. 
We introduce \textbf{AI-PAVE-Br}, an LLM-based system specifically engineered to navigate the unique linguistic and structural complexities of product descriptions in Portuguese. 
A cornerstone of our contribution is the \textbf{Golden Set}, a meticulously annotated and publicly shared dataset spanning 20 product types. 
This set stands as the first high-quality benchmark for PAVE in Portuguese, providing an invaluable and objective asset for validating our system and enabling future research.

Our experimental results are conclusive: \textbf{AI-PAVE-Br}, guided by targeted prompt engineering, dramatically outperforms traditional NER methods. The system excels not only at extracting core product entities but also demonstrates remarkable capability in identifying complex ``Category'' and ``Subcategory'' attributes directly from Portuguese product titles. This success underscores the transformative potential of LLMs to move beyond Anglocentric models and provide scalable, highly accurate solutions for a major global market. By delivering both a novel system and a foundational benchmark, our work paves the way for a new wave of research and development in e-commerce AI for the Portuguese-speaking world. 
Future work will focus on several key areas:

\begin{itemize}[noitemsep,topsep=0pt]
    \item \textbf{Adaptive Prompting:} Investigating dynamic prompt generation strategies or few-shot learning techniques to improve performance on long-tail product categories or newly emerging attributes without requiring extensive re-annotation.
    \item \textbf{Fine-tuned SLMs:} Exploring the performance of \textbf{AI-PAVE-Br} in comparison to fine-tuned SLMs, such as mBERT \cite{Devlin2019} and BERTimbau \cite{Souza2020}, in the context of Brazilian e-commerce.
    \item \textbf{Addressing Performance Dips:} Further analysis and targeted optimization for categories where LLM performance did not surpass or even fell below traditional baselines.
    \item \textbf{Output Normalization Challenges:} Developing robust methods for normalizing the extractor's output, as inconsistent formatting of extracted values remains a significant challenge for downstream applications.
\end{itemize}

Ultimately, the combination of high-quality golden data and advanced LLM techniques paves the way for a more efficient, accurate, and scalable product data infrastructure, critical for enhanced user experiences and operational efficiency in e-commerce.

\bibliographystyle{sbc}
\bibliography{references}

\end{document}